\pdfoutput=1
\documentclass[11pt]{article}

\usepackage[]{acl}

\usepackage{times}
\usepackage{latexsym}

\usepackage[T1]{fontenc}
\usepackage[utf8]{inputenc}

\usepackage{microtype}

\usepackage{amsmath,amsfonts,bm}

\def\va{{\bm{a}}}
\def\vb{{\bm{b}}}

\def\vh{{\bm{h}}}

\def\vx{{\bm{x}}}
\def\vy{{\bm{y}}}
\def\vz{{\bm{z}}}

\def\mB{{\bm{B}}}

\def\mW{{\bm{W}}}
\def\mX{{\bm{X}}}
\def\mY{{\bm{Y}}}

\DeclareMathAlphabet{\mathsfit}{\encodingdefault}{\sfdefault}{m}{sl}
\SetMathAlphabet{\mathsfit}{bold}{\encodingdefault}{\sfdefault}{bx}{n}

\usepackage{color}
\usepackage{xcolor}
\definecolor{ugreen}{rgb}{0,0.5,0}
\definecolor{lgreen}{rgb}{0.9,1,0.8}
\definecolor{lightgray}{gray}{0.85}
\definecolor{myblack}{rgb}{0.15,0.15,0.15}
\definecolor{lyellow}{rgb}{0.54, 0.25, 0.27}

\definecolor{darkblue}{rgb}{0.0, 0.0, 0.55}
\definecolor{darkcandyapplered}{rgb}{0.64, 0.0, 0.0}
\usepackage{caption}
\usepackage{enumitem}
\usepackage{amsmath}
\usepackage{amsfonts} % for font of real value symbol 'R'
\usepackage{pgfplots}
\usepackage{tikz}
\usepackage{subfig}
\usetikzlibrary{backgrounds,fit} % libarary for tikz
\usetikzlibrary{shapes,arrows,shadows}
\usetikzlibrary{patterns}
\usetikzlibrary{shapes.geometric} % for use 'triangle'
\usetikzlibrary{decorations.pathreplacing} % for use 'decorate brace'
\usetikzlibrary{calc}
\usepackage{array,multirow}
\usepackage{array} % set tabular width
\usepackage{booktabs} % use bold line in table
\usepackage{arydshln} % % draw dashed line in table
\usepackage{diagbox} % use diagbox in table 
\usepackage{bbm} % indicate function, e.g. 1
\newcommand{\PreserveBackslash}[1]{\let\temp=\\#1\let\\=\temp} % set table width
\newcolumntype{C}[1]{>{\PreserveBackslash\centering}p{#1}}
\newcolumntype{R}[1]{>{\PreserveBackslash\raggedleft}p{#1}}
\newcolumntype{L}[1]{>{\PreserveBackslash\raggedright}p{#1}}
\newcolumntype{M}[1]{ >{\centering\arraybackslash}m{#1}}
\usepackage{cleveref}
\usepackage{xspace}
\usepackage{algorithm,algorithmicx,algpseudocode}
\usepackage{multicol}
\usepackage{bm}

\pgfplotsset{compat=1.16}

\usepackage{verbatim}
\usepackage{float}

\newlength{\vseg}
\newlength{\hseg}
\newlength{\wnode}
\newlength{\hnode}
\newcommand{\knnmt}{\textit{k}NN-MT\xspace}
\newcommand{\clknn}{\textsc{Clknn}\xspace}

\title{Learning Decoupled Retrieval Representation for \\ Nearest Neighbour Neural Machine Translation}

\author{
	Qiang Wang$^{1,2}$,
	\textbf{Rongxiang Weng$^{3}$},
	Ming Chen$^2$\thanks{\xspace\xspace Corresponding author.} \\
	$^{1}$Zhejiang University, Hangzhou, China\\
	$^{2}$RoyalFlush AI Research Institute, Hangzhou, China\\
	$^{3}$School of Computer Science and Technology, Soochow University, Suzhou, China \\
	{\tt
		\{wangqiangenu, wengrongxiang\}@gmail.com, chenming@myhexin.com
	}\\
}

\begin{document}
\maketitle
\begin{abstract}

K-Nearest Neighbor Neural Machine Translation (\knnmt) successfully incorporates external corpus by retrieving word-level representations at test time. 
Generally, \knnmt borrows the off-the-shelf context representation in the translation task, e.g., the output of the last decoder layer, as the query vector of the retrieval task.
In this work, we highlight that coupling the representations of these two tasks is sub-optimal for fine-grained retrieval.
To alleviate it, we leverage supervised contrastive learning to learn the distinctive retrieval representation derived from the original context representation. We also propose a fast and effective approach to constructing hard negative samples. Experimental results on five domains show that our approach improves the retrieval accuracy and BLEU score compared to vanilla \knnmt.

\end{abstract}

\section{Introduction}

Conventional neural machine translation (NMT) cannot dynamically incorporate external corpus at inference once finishing training \cite{bahdanau2015neural,vaswani2017attention}, resulting in bad performance when facing unseen domains, even if feeding millions or billions of sentence pairs for training \cite{koehn2017six}.  
To address this problem, researchers developed retrieval-enhanced NMT (\textsc{Renmt}) to flexibly incorporate external translation knowledge. Early \textsc{Renmt}s leverage a search engine to find the similar bitext to improve the translation performance \cite{zhang-etal-2018-guiding,cao-xiong-2018-encoding,DBLP:conf/aaai/GuWCL18,DBLP:conf/aaai/XiaHLS19}. %However, in practical applications, sentences with high similarity are sparse and not always available. As a result, the translation performance may be damaged when sentences with low similarity are retrieved. \cite{cao-xiong-2018-encoding}.
However, the results of sentence-level retrieval with high similarity are generally sparse in practical applications, while noises in low similarity retrieval could lead to severe performance degradation \cite{cao-xiong-2018-encoding}.

\knnmt proposed by \citet{khandelwal2021nearest} effectively alleviates the sparse problem by introducing the word-level k-nearest neighbor mechanism. Instead of storing the discrete word sequence, \knnmt uses a pre-trained NMT model to force decoding the external corpus and remembers the word-level continuous context representation, e.g., the output of the last decoder layer. During inference, \knnmt assumes that the same target words have similar contextual representations and weights word selection through retrieving current context representation from the memorized datastore. However, we point out that it is sub-optimal to directly use the off-the-shelf context representation in the translation task because this vector is not specific to fine-grained retrieval.
 
In this work, we attempt to decouple the context representation by learning an independent retrieval representation. To this end, we leverage supervised contrastive learning with multiple positive and negative samples to learn a good retrieval representation (called \clknn). We also propose a fast and effective method to construct hard negative samples. Experimental results on five domains show that our approach outperforms the vanilla \knnmt in terms of BLEU and retrieval accuracy.

\section{Background}

\paragraph{Vanilla NMT} 
Given a source sentence $\vx=\{x_1, x_2, \ldots, x_{|\vx|}\}$ and a target prefix $\vy_{<t}=\{y_1, y_2, \ldots, y_{t-1}\}$, the vanilla NMT predicts the next target word $y_t$ by:
\begin{equation}
    \label{eq:p_c}
	p_{c}(y_t|\vx,\vy_{<t}) \propto \textrm{exp}\Big(q(\vh_{t})\Big)
\end{equation}
where $\vh_{t}=f_{\theta}(\vx, \vy_{<t}) \in \mathcal{R}^{d}$ is the context vector at step $t$ with respect to $\vx$ and $\vy_{<t}$; $f_{\theta}(\cdot)$ can be arbitrary encoder-decoder network with parameters $\theta$, such as Transformer \cite{vaswani2017attention}; $q(\cdot)$ linearly projects $\vh_{t}$ to target vocabulary size. 

\paragraph{\knnmt} 
\knnmt hypothesizes that the same target words have similar representations.
To dynamically incorporate external sentence pairs $\mathcal{D}=\{(\vx^{(i)}, \vy^{(i)})\}_{i=1}^{|\mathcal{D}|}$, \knnmt extends Eq.~\ref{eq:p_c} by interpolating a retrieval-based probability $p_{r}$:
\begin{equation}
    \label{eq:p_knn}
	p_{knn} = (1-\lambda) \times p_{c} + \lambda \times p_{r}
\end{equation}
where $\lambda$ is the interpolation coefficient as a hyper-parameter.

Specifically, \knnmt first uses a pre-trained NMT model to force decoding each sentence pair $(\vx^{(i)}, \vy^{(i)})$ to build a key-value datastore $\mathcal{H}$:
\begin{equation}
    \label{eq:knn_datastore}
    \mathcal{H} = \bigcup_{i=1}^{|\mathcal{D}|} \bigcup_{t=1}^{|\vy^{(i)}|} \Big\{(\vh^{(i)}_t, y_t^{(i)}) \Big\}
\end{equation}
The key is the word-level context representation $\vh^{(i)}_t$ and the value is the gold target word $y^{(i)}_t$.  
Then, given $\mathcal{H}$ and predicted target prefix $\hat{\vy}_{<t}$ at test time, \knnmt models $p_r(\hat{y_t}|\vx, \hat{\vy}_{<t})$ by measuring the distance between query $\hat{\vh}_t=f_{\theta}(\vx, \hat{\vy}_{<t})$ and its k-nearest representations $\{(\tilde{\vh}_i, \tilde{v}_i)\}_{i=1}^k$ in $\mathcal{H}$:
\begin{equation}
    \label{eq:knn_qr}
    p_r(\hat{y}_t|\vx, \hat{\vy}_{<t}) \propto \sum_{i=1}^k \mathbbm{1}_{\hat{y}_t=\tilde{v}_i}  \textrm{exp} \Big( \frac{-d(\tilde{\vh}_i, \hat{\vh}_t)}{T} \Big),
\end{equation}
where $d(\cdot)$ is $L_2$ distance; $T$ is temperature hyper-parameter; $\mathbbm{1}$ is the indicator function. 

\section{Approach}

%Algorithm~\ref{alg:training} summarizes the whole training process.
\paragraph{Motivation}
According to Eq.~\ref{eq:p_c}-\ref{eq:knn_qr}, we can see that the context representation $\vh$ simultaneously plays two roles in \knnmt: (1) the semantic vector for $p_c$; (2) the retrieval vector for $p_r$. 
We note that coupling the same $\vh$ in the two scenes is sub-optimal. 
Recall that $\vh$ in the translation model is generally learned through cross-entropy loss, which only pays attention to the gold target token and ignores others.\footnote{In practice, we often use its label-smooth variant, which evenly assigns a small probability mass to all non-gold labels without distinction.} 
However, a good retrieval vector should be able to distinguish between different tokens, especially those owning similar representations. 
Therefore, we attempt to derive a new retrieval vector $\vz$ from $\vh$ for better retrieval performance.

\paragraph{Retrieval representation adapter}

\iffalse
\begin{algorithm}[tb]
\caption{Training Algorithm for \clknn}  
\label{alg:training}  
\begin{algorithmic}[1]
	\Require Training data $D$ including distillation targets, pretrained AT model $\textrm{M}_{at}$, chunk size $k$, mixed distillation rate $p_{raw}$, schedule coefficient $\lambda$
	\Ensure Hybrid-Regressive Translation model $\textrm{M}_{hrt}$
	
	\State $\textrm{M}_{hrt} \gets \textrm{M}_{at}$ \Comment{fine-tune on  pre-trained AT}
	\For {$t$ in $1,2,\ldots,T$} 
	\State   $\mX=\{\vx_1, \ldots, \vx_n\}$, $\mY=\{\vy_1, \ldots, \vy_n\}$, $\mY'=\{\vy'_1, \ldots, \vy'_n\}$ $\gets$ fetch a batch from $D$
	
	\For {$i$ in $1,2,\ldots,n$} 
	\State   $\mB_i=(\mX_i, \mY^*_i) \gets$ 
	sampling $\mY^*_i$ $\sim$ \{$\mY_i$, $\mY'_i$\} with $P(\mY_i)=p_{raw}$  \Comment{mixed distillation} 
	\EndFor
	
	\State   $p_{k} \gets (\frac{t}{T})^\lambda $ %get the chunk-aware proportion by Eq.~\ref{eq:skip_rate} 
	\Comment{curriculum learning}
	
	\State   $\mB_{c=k}, \mB_{c=1} \gets \mB_{:\lfloor n \times p_k \rfloor}, \mB_{\lfloor n \times p_k \rfloor:} $ \Comment{split batch}
	
	\State   $\mB_{c=k}^{at}, \mB_{c=k}^{nat} \gets$ construct \{Skip-AT, Skip-CMLM\} training samples based on $\mB_{c=k}$ 
	\State   $\mB_{c=1}^{at}, \mB_{c=1}^{nat} \gets$ construct \{AT, CMLM\} training samples based on $\mB_{c=1}$
	
	%\State   $\mathcal{B} \gets 
	\State   Optimize $\textrm{M}_{hrt}$ using $\mB_{c=k}^{at} \cup \mB_{c=1}^{at} \cup \mB_{c=k}^{nat} \cup \mB_{c=1}^{nat}$ \Comment{joint training} 
	\EndFor
\end{algorithmic}
\end{algorithm}
\fi

We use a simple feedforward network as an adapter to transform the original representation $\vh$ to desired retrieval representation $\vz$:
\begin{equation}
    \label{eq:ffn}
    \vz = \textrm{FFN}(\vh) =  \textrm{ReLU}(\vh \mW_1 + \vb_1)\mW_2 + \vb_2, 
\end{equation}
where $\mW_1 \in \mathcal{R}^{d \times d_f}$, $\mW_2 \in \mathcal{R}^{d_f \times d_o}$, $\vb_1 \in \mathcal{R}^{d_f}$, and $\vb_2 \in \mathcal{R}^{d_o}$ are learnable parameters; $d_f$ and $d_o$ are the intermediate hidden size and output size of the adapter, respectively. When $d_o < d$, the adapter network can be regarded as a dimension reducer. As \textrm{FFN} is very lightweight compared to the calculation of $\vh$, there is almost no latency 
in converting $\vh$ to $\vz$.%\footnote{The inference speed of CLKNN is comparable to that of knn-MT (x0.97 +- 0.02) measured on IT test set in five runs. The reason is that our only newly add one FNN to construct the retrivial representation, whose input/hidden/output dimensions are 1024/4096/512, respectively. In contrast, the original model composes of 6 larger FNN with the dimensions of 1024/4096/1024. We will add these results in the revised version.}
For convenience, in the following description, we redefine $\vh_i$ as the key of $i$-th key-value pair in the original datastore $\mathcal{H}$, and the corresponding value is denoted by $Y_i$ when there is no ambiguity.
In this way, the new datastore $\mathcal{Z}$ can be denoted as $\mathcal{Z}=\{(\vz_i, Y_i)| i=1,\ldots,|\mathcal{H}|\}$, where $\vz_i=\textrm{FFN}(\vh_i)$.

\paragraph{Supervised contrastive learning}
In machine translation field, contrastive learning has been applied in multilingual translation \cite{pan-etal-2021-contrastive,wei2021on}, cross-modal translation \cite{ye-etal-2022-cross}, and learning robust representation for low-frequency word \cite{zhang-freq-aware} etc.
In this work, we use supervised contrastive learning \cite{NEURIPS2020_f3ada80d} with multiple positive and negative samples to learn the desired retrieval representation $\vz$. Here, we regard the unique token $v$ in the target vocabulary $V$ as a natural supervision signal. We aim to make $\vz$ more distinguishable, for example, pulling $\vz$ of the same words together and pushing $\vz$ of different words apart.
Specifically, we first divide $\mathcal{Z}$ into $|V|$ clusters according to the token class label. E.g., $C_v = \{\vz_i| i=1,\ldots,|\mathcal{Z}|, Y_i=v\}$, where $C_v$ is the context representation cluster of token $v$. Thus, given any context representation $\vz \in \mathcal{Z}$ and its token label $v$, we can construct M positive samples $\vz^+=\{\vz_1^+, \ldots, \vz_i^+, \ldots, \vz_M^+\}$, where $\vz_i^+$ is uniformly sampled from its owned cluster $C_v$ and $\vz_i^+ \ne \vz$.\footnote{We use sampling with replacement when $|C_v| < M$.} Likely, we further construct N negative samples $\vz^-=\{\vz_1^-, \ldots, \vz_i^-, \ldots, \vz_N^-\}$, where $\vz_i^- \in \backslash C_v$, $\backslash C_v$ denotes other clusters except $C_v$. In the next part, we will describe how to build $\vz^-$.
%Then we leverage this label information to construct a cluster by collecting all its representations: $C_v = \{(\textrm{FFN}(\vh_i), y_i), i=1,\ldots,|\mathcal{H}|, y_i=v\}$. 
%Thus, supervised contrastive learning enables to (1) pull the points belonging to the same cluster together; (2) push apart points in different clusters.
%Specifically, given an anchor vector $\vz$ and its label $v$, we learn the adapter network by the following training objective:
Finally, given the anchor vector $\vz$, its multiple positive samples $\vz^+$ and multiple negative samples $\vz^-$, we learn the adapter network through the following contrastive learning loss:
\begin{equation}
    \label{eq:infonce}
    - \textrm{log} \frac{\sum\limits_{1 \le i \le M}\textrm{exp}(s(\vz, \vz_{i}^+))}{\sum\limits_{1 \le i \le M}\textrm{exp}(s(\vz, \vz_{i}^+)) + \sum\limits_{1 \le j \le N}\textrm{exp}(s(\vz, \vz_{j}^-))},
\end{equation}
where 
%M, N is the number of positive and negative samples, respectively; $\vz^+$, $\vz^-$ is the set of positive samples and negative samples about $\vz$, respectively; 
$s(\cdot)$ is the score function implemented as cosine similarity with temperature $T'$: $s(\va,\vb)=\frac{1}{T'} \times \frac{\va^T \vb}{\Vert \va \Vert \cdot \Vert \vb \Vert}$.
Note that $T'$ is the temperature in training, which is different from the inference temperature $T$ in Eq.~\ref{eq:knn_qr}.

\begin{table*}[htb]
	\begin{center}
		
	\begin{tabular}{l c c c c c c}
		\toprule[1pt]

        \multicolumn{1}{c}{\textbf{Dataset}} & \multicolumn{1}{c}{\textbf{Medical}} &
		\multicolumn{1}{c}{\textbf{Law}} &
        \multicolumn{1}{c}{\textbf{IT}} &
		\multicolumn{1}{c}{\textbf{Koran}} &
		\multicolumn{1}{c}{\textbf{Subtitle}} &
		\multicolumn{1}{c}{\textbf{NC+Euro}} \\
		\hline %\hline

		Train & 248K & 467K & 222K & 52K & 500K & 2M \\
		Valid & 2000 & 2000 & 2000 & 2000 & 2000 & -\\
		Test & 2000 & 2000 & 2000 & 2000 & 2000 & -\\ \hline
		Datastore & 6.9M & 19.0M & 3.6M & 0.5M & 6.2M & 5M$^\dagger$ \\ 
			
	\bottomrule[1pt]
	\end{tabular}
		
	%\vspace{-.5em}
	\caption{Statistics of datasets in different domains. $\dagger$: Due to limited memory, we randomly sampled 5M samples from a total of 65.7M samples in NC+Euro for training.}
	\label{table:data}
	%\vspace{-1em}
	\end{center}
\end{table*}

\begin{table*}[t]
	\begin{center}
	\resizebox{0.8\textwidth}{!}
	{
	\begin{tabular}{l c c c c c c c}
		\toprule[1pt]

        \multicolumn{1}{c}{\textbf{Method}} & \multicolumn{1}{c}{\textbf{Medical}} &
		\multicolumn{1}{c}{\textbf{Law}} &
        \multicolumn{1}{c}{\textbf{IT}} &
		\multicolumn{1}{c}{\textbf{Koran}} &
		\multicolumn{1}{c}{\textbf{Subtitle}} &
		\multicolumn{1}{c}{\textbf{Avg.}}
		\\
		\hline %\hline
				
		Baseline (WMT19 winner, \citet{ng-etal-2019-facebook}) & 39.91 & 45.71 & 37.98 & 16.3 & 29.21 & 33.82 \\
				
		\knnmt \cite{khandelwal2021nearest} & 54.35 & 61.78 &	45.82 &	19.45 & \textbf{31.73}$^\dagger$ & 42.63 \\
				
		\knnmt (our implementation)	&	54.41 &	61.01 &	45.20 & 21.07 & 29.67 & 42.27 \\
		\hline
		 
		% learn representation by out-domain
		\multicolumn{7}{c}{\textit{train by out-domain data}} \\
		\hline
		\clknn & 56.37 &	61.54 &
46.50 &	21.52 &	30.81 & 43.35 \\
				
		\clknn + $\lambda^*$ & \textbf{56.52} &	61.63 & 46.68 &	21.60 &	30.86 & 43.46 \\
		\hline
		\multicolumn{7}{c}{\textit{train by in-domain data}} \\
		\hline
		% learn repsentation by in-domain
		\clknn & 55.86 & 61.92 &	47.77 & 21.46 & 31.02 & 43.61 \\
			
		\clknn + $\lambda^*$ & 55.87 & \textbf{62.01} & \textbf{47.84} & \textbf{21.81} & 31.05 & \textbf{43.72} \\
			
	\bottomrule[1pt]
	\end{tabular}
	}
		
	\vspace{-.5em}
	\caption{The SacreBLEU scores of our proposed \clknn and the baseline methods in five domains. $\lambda^*$ denotes using retrieval confidence aware interpolation coefficient. $^\dagger$ denotes the number is not comparable because \citet{khandelwal2021nearest} use full-size subtitle data than ours.
	All the \clknn results are significantly better (p$<$0.01) than our re-implemented \knnmt, measured by paired bootstrap resampling \cite{koehn-2004-statistical}.}
	\label{table:main_results}
	\vspace{-.5em}
	\end{center}
\end{table*}

\paragraph{Fast hard negative sample}
%For eq.~\ref{eq:infonce}, the positive samples $\vz^+$ are easy to collect: We uniformly random sampling from its cluster $C_v$. However, it is a challenge to construct good negative samples $\vz^-$.
The key for Eq.~\ref{eq:infonce} is the construction of negative samples $\vz^-$. 
A trivial solution is randomly sampling from the entire space of $\backslash C_v$. However, this negative sample may be too easy to provide the effective learning signal \cite{robinson2020contrastive}. On the contrary, an extreme method for hard negative samples is to traverse $\backslash C_v$ to find the most similar negative samples for the anchor.  
The problem is that $|\backslash C_v|$ is close to $|\mathcal{Z}|$, with a scale of millions or more, resulting in enormous computational complexity.
To solve it, we propose a fast and cheap approach to constructing hard negative samples. Specifically, we first collect the cluster centre $\bar{C}_v = \frac{1}{|C_v|} \sum_{i=1}^{|\mathcal{Z}|} \mathbbm{1}_{Y_i=v} \vz_i$. We calculate the nearest K (K$>=$N) cluster centers w.r.t the anchor and randomly sample N clusters to make the source of the negative sample diverse. Then we randomly sample one point from the corresponding cluster as a negative sample. As the anchor vector only involves querying $|C|$ cluster centers and $|C| << |\mathcal{Z}|$, our approach runs faster than the exact global search.

\paragraph{Inference}
After training, we use the well-trained $\textrm{FFN}$ to rebuild the retrieval datastore $\mathcal{H}$ into $\mathcal{Z}$. To further reduce calculation cost at test time, we introduce PCA to reduce the dimension of the retrieval vector. We also add normalization after PCA to guarantee the numerical stability of the input to the inner product. Another difference with Eq.~\ref{eq:knn_qr} is that we use the inner product instead of the L2 distance as distance metrics. The reason is that using consistent distance metrics in training and inference improves performance in primitive experiments.
Concretely, we modify the original \knnmt in Eq.~\ref{eq:knn_qr} as:
\begin{equation}
    \label{eq:new_qr}
    p_r(\hat{\vy}_t|\vx, \hat{\vy}_{<t}) \propto \sum_{i=1}^k \mathbbm{1}_{\hat{y}_t=\tilde{v}_i}  \textrm{exp} \Big( \frac{g(\tilde{\vz}_i) \otimes g(\hat{\vz}_t)}{T} \Big),
\end{equation}
where $g(x) = \textrm{Norm}(\textrm{PCA}(x))$, $\otimes$ denotes inner product operation, $\tilde{\vz}_i$ is the $i$-th nearest neighbor in $\mathcal{Z}$ for the current retrieval representation $\hat{\vz}_t$.
As a bonus, since the numeric range of the normalized inner product is $[0, 1]$, which can be seen as the confidence in retrieving.\footnote{L2 distance lacks this feature because its numeric range is too broad, e.g., {0\textasciitilde1000} in our observation.} We leverage this nature to modify the interpolation coefficient $\lambda$ in Eq.~\ref{eq:p_knn} to be aware of retrieval confidence:
\begin{equation}
    \label{eq:new_qr}
    \lambda^* = \lambda \times \frac{\sum_{i=1}^k g(\tilde{\vz}_i) \otimes g(\hat{\vz}_t)}{k}.
\end{equation}
%where $\tilde{\vz}$ is the nearest neighbour w.r.t $\vz$.
$\lambda^*$ can be considered a simple adaptive coefficient like \citet{zheng-etal-2021-adaptive,jiang-etal-2021-learning,wang-cluster-knn}, but does not require training.

%%%%%%%%%%%%%%%%
%% FIGURE STARTS
%% FIGURE: training samples
\begin{figure}[t]
	\begin{center}
	%\resizebox{\textwidth}{!}
	%{
	\includegraphics[width=.48\textwidth]{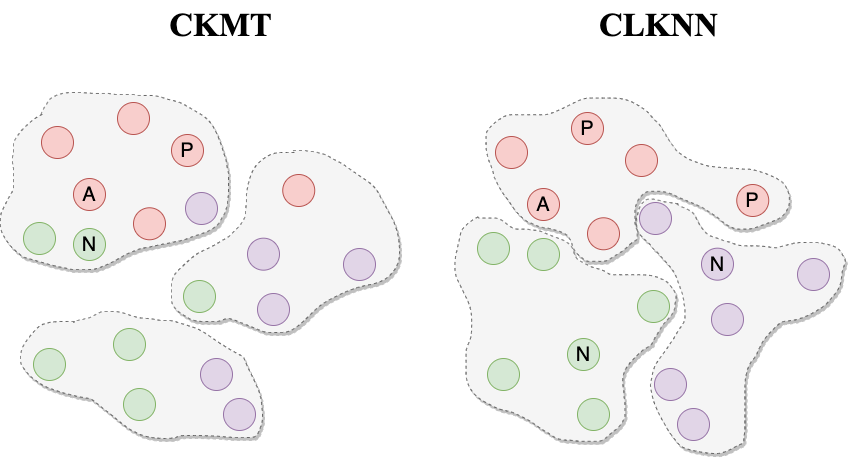}
	%	}
	\end{center}
	
	\begin{center}
		\vspace{-.5em}
		\caption{Illustration the differences between CKMT and \clknn in constructing positive and negative samples. Different colors indicate different tokens. \texttt{A}/\texttt{P}/\texttt{N} means anchor, positive sample and negative sample, respectively.  }
	\label{fig:difference}
		\vspace{-1.2em}
	\end{center}
\end{figure}

\paragraph{Discussion}
The closest work with us is CKMT \cite{wang-cluster-knn}. As illustrated in Figure~\ref{fig:difference}, there are two major differences compared with CKMT:
(1) \clknn uses multiple positive and negative samples, while CKMT only considers a single positive and negative sample, limiting the exploration of representation space.
(2) CKMT requires to partition clusters through cost-expensive clustering in full-scale datastore, while \clknn predefines clusters based on vocabulary labels and only involves calculating cluster centers. 
In practice, we spent about 6 hours on the CPU to complete the cluster operation in CKMT, while \clknn only takes about 3 minutes. 

\section{Experiments}
\label{sec:exp}

%To validate the effectiveness of CLKNN, we conducted experiments on the domain adaptation scene.

%\subsection{Setup}
\paragraph{Setup}

%%%%%%%%%%%%%%%%
%% FIGURE STARTS
%% FIGURE: training samples
\begin{figure*}[t]
	\begin{center}
	\resizebox{0.8\textwidth}{!}
	{
	
	\begin{tabular}{C{.3\textwidth}C{.3\textwidth}C{.3\textwidth}}
		\subfloat [\small{High}] {
	    	\includegraphics[width=.3\textwidth]{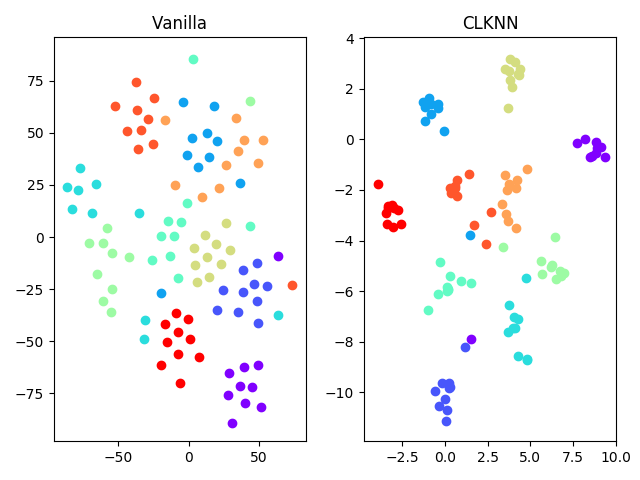}
		}
		&
		\subfloat [\small{Middle}
		] {
			\includegraphics[width=.3\textwidth]{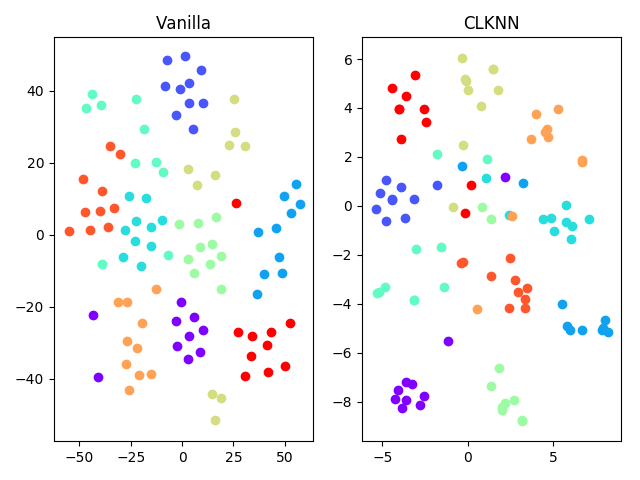}
		}
		&
		\subfloat [\small{Low}
		] {
			\includegraphics[width=.3\textwidth]{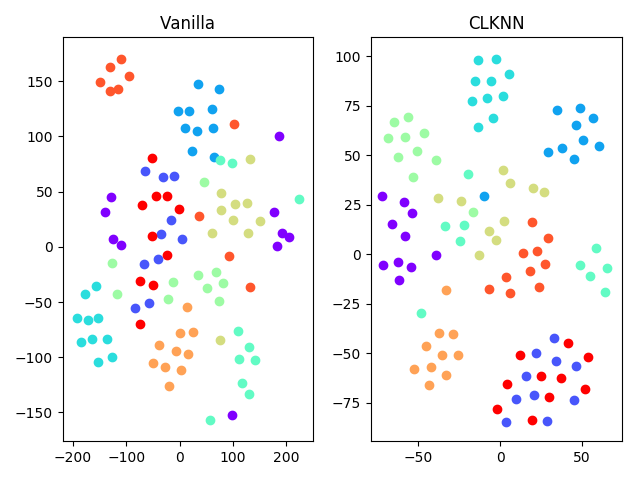}
		}
		\\
		\end{tabular}
		}
	\end{center}
	
	\begin{center}
		\vspace{-.5em}
		\caption{Visualization of retrieval vector on different frequency words by t-SNE. We uniformly sample 10 classes in each category, and each class contains ten random representations. The same color denotes the same class.  }
	\label{fig:visual}
		\vspace{-1.2em}
	\end{center}
\end{figure*}

\noindent 
To fairly compared with previous work \cite{khandelwal2021nearest}, 
we use WMT'19 German-English news translation task winner \cite{ng-etal-2019-facebook} as our strong general domain baseline.
We use the same German-English multi-domain datasets, consisting of five domains, including \texttt{Medical}, \texttt{Law}, \texttt{IT}, \texttt{Koran} and \texttt{Subtitles} \footnote{We use the provided 500K sentence pairs version subtitle data rather than full size 12.4M due to memory limitation.}. Besides, to test the proposed training approach robust in out-domain scenery, we also use a 2M subset of the baseline's training data, including \textit{News Commentary v14} and \textit{Europarl v9}, and randomly sample 5M samples out of 65.7M samples from its datastore. See Table~\ref{table:data} for detailed data statistics.

%\subsection{Implementation details}
\paragraph{Implementation details}

\noindent 
All experiments run on a single NVIDIA 2080 Ti GPU. We use \textit{Faiss} \footnote{\url{https://github.com/facebookresearch/faiss}} for vector retrieval. For \clknn, the number of positive samples is M=2, and the number of negative samples is N=32. We sample N negative samples from K=128 nearest clusters. The training batch size is 32. During training, we set $T^{'}$=0.01, while we vary $T$ according to the validation set at test time. The hidden state size $d_f$ and output size $d_o$ of adapter is 4096 and 512, respectively. The output dimension of PCA is 128.
We train all models for 500k steps and select the best model on the validation set. We use a beam size of 5 and a length penalty of 1.0 for all experiments for inference.  
We measure case-sensitive detokenized BLEU by SacreBLEU.

%\subsection{Experimental results}
\paragraph{Experimental results}

Table~\ref{table:main_results} reports the SacreBLEU scores in five domains. We can see that: (1) {\clknn} is robust about training data: using out-domain or in-domain average improves 1+ points than our \knnmt; (2) The gap between in-domain and out-domain is small (about 0.3 points), meaning that our approach does not rely on in-domain data and is more practical than \citet{zheng-etal-2021-adaptive,jiang-etal-2021-learning}; (3) using proposed $\lambda^*$ slightly improve the performance across the board. These results show that learning independent retrieval representation is helpful for vanilla \knnmt. Besides, we also compare the inference speed between \clknn and \knnmt through running five times on \texttt{IT} test set. The results show that \clknn has a comparable speed (97\%$\pm$2\%) to that of \knnmt because the adapter in \clknn is very lightweight.

\section{Analysis}
\label{sec:analysis}

%\subsection{Effect of the number of contrastive sample}
\paragraph{Effect of the number of contrastive samples}

\begin{table}[t]
	\begin{center}
	\resizebox{0.35\textwidth}{!}
	{
	\begin{tabular}{c c c | c c c}
		\toprule[1pt]

        \multicolumn{1}{c}{\textbf{M}} & \multicolumn{1}{c}{\textbf{N}} &
		\multicolumn{1}{c|}{\textbf{BLEU}} &
		\multicolumn{1}{c}{\textbf{M}} & \multicolumn{1}{c}{\textbf{N}} &
		\multicolumn{1}{c}{\textbf{BLEU}} \\
		\hline %\hline

		1 & 1 & 45.54 &  2 & 16 & 46.37\\
		1 & 16 & 45.91 & 2 & 32 & 46.68\\
		1 & 32 & 46.13 & 2 & 64 & 46.55 \\
		1 & 64 & 45.88 & 4 & 32 & 46.29\\

	\bottomrule[1pt]
	\end{tabular}
	}
	\vspace{-.5em}
	\caption{The BLEU scores on \texttt{IT} test set against the number of the positive (M) and negative (N) samples.}
	\label{table:sample_num}
	\vspace{-1em}
	\end{center}
\end{table}

One of the main differences between \citet{wang-cluster-knn} and us is that we use multiple positive and negative samples in our training objective. We vary the number of M and N and report the BLEU scores in Table~\ref{table:sample_num}. As we can see, increasing M and N is helpful for our method. However, large M cannot befit more than increasing N. We attribute it to positive samples that are too easy to learn because most are close in embedding space.
On the contrary, negative samples from different clusters can provide a stronger learning signal. To further validate the effectiveness of multiple samples, we also conduct experiments on \texttt{Medical}. The results are similar to that of \texttt{IT}: using M=2, N=32 is 1.64 BLEU points higher than using M=1, N=1 (56.52 vs. 54.88). It indicates that using multiple positive and negative samples is necessary to achieve good performance for contrastive learning.

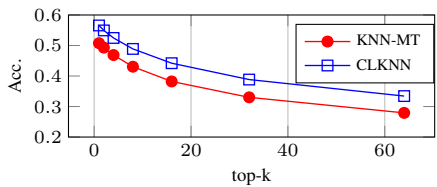
\begin{figure}[tbp]
	\begin{center}
		\renewcommand\arraystretch{0.0}
		\setlength{\tabcolsep}{1pt}
		\begin{tabular}{c}

				\begin{tikzpicture}{baseline}
				\scriptsize{
					\begin{axis}[
					xmajorgrids,
					ylabel near ticks,
					width=.4\textwidth,
					height=.2\textwidth,
					legend style={at={(0.62,0.7)}, anchor=south west},
					xlabel={\scriptsize{top-k}},
					ylabel={\scriptsize{Acc.}},
					ylabel style={yshift=-0em},xlabel style={yshift=0.0em},
					ymin=0.2,ymax=0.6,%ytick={49.0, 49.5, 50.0, 50.5, 51.0, 51.5, 52.0, 52.5},
					legend style={yshift=-12pt, legend plot pos=left,font=\tiny,cells={anchor=west}}
					]
					
					\addplot[red,mark=otimes*,line width=0.5pt] coordinates {(1, 0.5076) (2, 0.4931) (4, 0.4687) (8, 0.4305) (16,0.3822) (32,0.3301) (64,0.2792)};
						\addlegendentry{KNN-MT}
					%\addlegendentry{pre-norm}
					\addplot[blue,mark=square,line width=0.5pt] coordinates {(1, 0.5657) (2, 0.5494) (4, 0.5247) (8, 0.4888) (16,0.4420) (32,0.3884) (64,0.3343)};
						\addlegendentry{CLKNN}
					%\addlegendentry{\plainname}

					\end{axis}
				}
				\label{fig:enc_depth_zh2en}
				\end{tikzpicture}
		%	}
			
		\end{tabular}
	\end{center}
	
	\begin{center}
		\vspace{-0.5em}
		\caption{Retrieval accuracy curve against top-k.}
		\label{fig:acc}
		\vspace{-1.0em}
	\end{center}
\end{figure}

%\subsection{Retrieval accuracy}
\paragraph{Retrieval accuracy}
Intuitively, our approach can learn more accurate retrieval representation than vanilla \knnmt. To validate this hypothesis, we use \texttt{IT} validation as the datastore and plot the retrieval accuracy on top-k in Figure~\ref{fig:acc}. We can see that {\clknn} has more robust retrieval accuracy than \knnmt no matter how k changes. It indicates that the performance improvement comes from our better retrieval representation.

%\subsection{Virsulization}
\paragraph{Visualization}

We visually present the differences between baseline and {\clknn} on embedding space. Specifically, we split three categories according to the word frequency in \texttt{IT} training set: \texttt{HIGH}(the first 1\%), \texttt{Middle}(40\%-60\%) and \texttt{LOW}(the last 1\%) \footnote{We filter words whose frequency is less than 10.}. We uniformly sample 10 unique words in each category and randomly sample 10 unique vector representations from the training datastore. We use t-SNE to plot these representations, as illustrated in Figure~\ref{fig:visual}. We can see that: (1) high-frequency words' representations are prone to distinguish for both baseline and {\clknn}; (2) {\clknn} has more close distances in the same vocabulary than baseline; (3) {\clknn} has more robust accuracy for low-frequency words. %It indicates that CLKNN is more suitable for vector retrieval than the original representation.

\iffalse
\section{Related work}
\label{sec:related_work}

\paragraph{Retrieval augmented NMT}

Earlier retrieval-augmentated NMT mainly focus on sentence-level retrieval on bitext \cite{cao-xiong-2018-encoding,zhang-etal-2018-guiding,DBLP:conf/aaai/GuWCL18,DBLP:conf/aaai/XiaHLS19}. By leveraging the search results from the out-of-shefle search engine, \citet{cao-xiong-2018-encoding,DBLP:conf/aaai/GuWCL18} modify the network architecture to adapt for multi-source input and \citet{DBLP:conf/aaai/XiaHLS19} model the datastore via graph network. Instead of changing the original network architecture,  \citet{zhang-etal-2018-guiding} reward the model score when target n-grams of similar sentences appear on translation hypothesis; \citet{xu-etal-2020-boosting,bulte-tezcan-2019-neural} leverage data argumentation by similar sentence pairs. In addition, \citet{cai-etal-2021-neural} extend to retrieve on target monolingual data. 

\paragraph{Contrastive learning}
Contrastive learning is an effective unsupervised representation method, which is widely used in computer vision~\cite{pmlr-v119-chen20j,NEURIPS2020_f3ada80d,DBLP:conf/icml/ZbontarJMLD21}, natural language process~\cite{reimers-gurevych-2019-sentence} and cross-modal~\cite{pmlr-v139-radford21a}. 
In machine translation field, contrastive learning has been applied in multilingual NMT \cite{pan-etal-2021-contrastive,wei2021on}, learning robust representation for low-frequency word \cite{zhang-freq-aware} etc. The closest work with us is CKMT \cite{wang-cluster-knn}.

\fi

\section{Conclusion}
\label{sec:conclusion}

In this work, we proposed to use supervised contrastive learning to decouple the context representation from vanilla \knnmt. Experimental results on several tasks show that our approach outperforms \knnmt and learns a more accurate retrieval representation.

\section*{Acknowledgements}

We would like to thank the anonymous reviewers for the helpful comments. We also thank Shuqin Pan for the writing suggestions. %{\color{red}{This work is supported by the National Key R&D Program of China under Grant No.2017YFB0803301 and No. 2018YFB1403202.} }

% Entries for the entire Anthology, followed by custom entries
\bibliography{anthology,custom}
\bibliographystyle{acl_natbib}

\iffalse
\appendix

\section{Appendix A. Data Statistics}
\label{sec:appendix}

\begin{table*}[htb]
	\begin{center}
		
	\begin{tabular}{l c c c c c c}
		\toprule[1pt]

        \multicolumn{1}{c}{\textbf{Dataset}} & \multicolumn{1}{c}{\textbf{Medical}} &
		\multicolumn{1}{c}{\textbf{Law}} &
        \multicolumn{1}{c}{\textbf{IT}} &
		\multicolumn{1}{c}{\textbf{Koran}} &
		\multicolumn{1}{c}{\textbf{Subtitle}} &
		\multicolumn{1}{c}{\textbf{NC+Euro}} \\
		\hline %\hline

		Train & 248K & 467K & 222K & 52K & 500K & 2M \\
		Valid & 2000 & 2000 & 2000 & 2000 & 2000 & -\\
		Test & 2000 & 2000 & 2000 & 2000 & 2000 & -\\ \hline
		Datastore & 6.9M & 19.0M & 3.6M & 0.5M & 6.2M & 5M \\ 
			
	\bottomrule[1pt]
	\end{tabular}
		
	%\vspace{-.5em}
	\caption{Statistics of datasets in different domains.}
	\label{table:data}
	%\vspace{-1em}
	\end{center}
\end{table*}

%This is an appendix.
We list the statistics of used datasets in Table~\ref{table:data}. For \texttt{Subtitle}, we use the provided 500K sentence pairs version subtitle data rather than full size 12.4M due to memory limitation. Likely, we do not use the whole samples in \texttt{NC+Euro} dataset: We randomly sample 5M samples out of 65.7M samples.
\fi

\end{document}